\documentclass[review]{elsarticle}

\usepackage{lineno,hyperref}

\usepackage{graphicx} 
\usepackage{amsmath}
\usepackage{amsfonts}
\usepackage{hyperref}
\usepackage{enumitem}
\usepackage[export]{adjustbox}
\usepackage{fancyhdr}
\usepackage{lastpage}
\usepackage{relsize} 
\usepackage{microtype}

\newcommand{\dia}{\mathlarger{\mathlarger{\mathlarger{\diamond}}}}
\usepackage{color}

\begin{document}

\begin{frontmatter}

\title{Grounding Value Alignment with Ethical Principles
%\tnoteref{mytitlenote}
}
%\tnotetext[mytitlenote]{footnote}


\author[CMU]{Tae Wan Kim}
\ead{twkim@andrew.cmu.edu}

\author[Penn]{Thomas Donaldson}
\ead{donaldst@wharton.upenn.edu}

\author[CMU]{John Hooker}
\ead{jh38@andrew.cmu.edu}

\address[CMU]{Carnegie Mellon University, USA}
\address[Penn]{University of Pennsylvania, USA}

\begin{abstract}
An important step in the development of value alignment (VA) systems in AI is understanding how values can interrelate with facts.  Designers of future VA systems will need to utilize a hybrid approach in which ethical reasoning and empirical observation interrelate successfully in machine behavior.  In this article we identify two problems about this interrelation that have been overlooked by AI discussants and designers.  The first problem is that many AI designers commit inadvertently a version of what has been called by moral philosophers the ``naturalistic fallacy,'' that is, they attempt to derive an ``ought'' from an ``is.''  We illustrate when and why this occurs.  The second problem is that AI designers adopt training routines that fail fully to simulate human ethical reasoning in the integration of ethical principles and facts.  Using concepts of quantified modal logic, we proceed to offer an approach that promises to simulate ethical reasoning in humans by connecting ethical principles on the one hand and propositions about states of affairs on the other.

\end{abstract}


\end{frontmatter}

%\linenumbers

Declarations of interest: none

\section{Introduction}

Artificial intelligence (AI) is an attempt to imitate human intelligence. Indeed, Alan Turing's idea of the Turing Test was originated from ``the Imitation Game''---a game in which a man imitates a woman's behavior to deceive the interrogator sitting in a different room \cite{turing1950imitation}.  Research shows that AI has imitated much of human intelligence, especially, calculative and strategic intelligence (e.g., AlphaGo's victory over human champions). As highly developed AI technologies are rapidly adopted for automating decisions, societal worries about the compatibility of AI and human values are growing.  As a response, researchers have examined how to imitate moral intelligence as well as calculative and strategic intelligence. Such attempts are lumped under the broader term, ``value alignment'' (hereafter VA). Given that a unique element of human intelligence is moral intelligence---a capability to make an ethical decision--the attempt to imitate moral promises to bring AI to another, higher level. 

Russell et al.\ \cite{russell2015research} highlighted the need for VA and identified options for achieving it:

\begin{quote}
    ``[A]ligning the values of powerful AI systems with our own values and preferences ... [could involve either] a system [that] infers the preferences of another rational or nearly rational actor by observing its behavior ... [or] could be explicitly inspired by the way humans acquire ethical values.'' 
\end{quote} 

\noindent
This passage reflects two types of goals that are frequently attributed to VA.  One is to teach machines human preferences, and another is to teach machines ethics.  The word ``values'' in fact has this double meaning.  It can refer to what humans value in the sense of what they see as desirable, or it can refer to ethical principles.  The distinction is important, because we acquire knowledge of the two types of values in different ways.  

Values in the first sense are phenomena that can be studied empirically by observing human activity.  That is to say, values in this sense can be studied using observation or  experience to determine facts about human behavior, in particular, about what humans actually value.  Values in the second sense cannot be inferred solely from observation or experience, and an attempt to do so commits the {\em naturalistic fallacy}.  This is the fallacy of inferring what is right solely from a description of a state of affairs, or in short, inferring an ``ought'' from an ``is.''  For example, the fact that people engage in a certain activity or believe that it is ethical, does not in itself imply that the activity is ethical.  Thus if VA is based solely on inverse reinforcement learning or other empirical methods, it cannot yield genuine ethical principles.  Observation must somehow be combined with ethical principles that are obtained non-empirically.  We illustrate some failures to do so that either commit the naturalistic fallacy (of deriving an ``ought'' from an ``is'') or oversimplify the process of moral reasoning.  Finally, we move to suggest a hybrid process that interrelates values and facts using concepts of quantified modal logic.

Our proposal differs from the hybrid approach of Allen et al.\ \cite{AllenSmitWallach2005}, who recommend combining {\em top-down} and {\em bottom-up} approaches to machine ethics.  The top-down approach installs ethical principles directly into the machine, while the bottom-up approach asks the machine to learn ethics from experience.  The distinction may seem related to ours but is quite different.  Both  approaches combine the ideas of top-down and bottom-up, but Allen et al.\ focus on pedagogy whereas we focus on epistemology.  We confront, as Allen et al.\ do not, the issue of the proper justification of ethical behavior.  Allen et al.\ are concerned with the process of teaching machines to internalize ethics, whereas our approach raises the question of what counts as ethical reasoning.  From an epistemological perspective, Allen et al.'s  bottom-up approach can result in teaching strategies that either conflate ``is''s and ``ought''s or that effect their separation \emph{ad hoc}.  They suggest, for example, that a machine might learn ethics through a simulated process of evolution.  This commits the naturalistic fallacy, because the fact that certain ethical norms evolve does not imply that they are valid ethical principles.  Elsewhere they suggest that a bottom-up teaching style might also ``seek to provide environments in which appropriate behavior is selected or rewarded.'' Adopting this style, trainers could reward behavior that is viewed {\em a priori} to be right, which is a reward strategy that parents successfully use with children.  Yet while such a hybrid approach has practical advantages as a teaching strategy, it does not add up to genuine ethical reasoning. 

Another version of a hybrid approach to VA is advocated by Arnold et al.\ \cite{arnold2017value}, who suggest that ethical rules can be imposed as constraints on what a machine learns from observation.  Their motivation is to ensure that the machine is not influenced by bad behavior.  Yet the epistemic question cannot be avoided. We must ask what remains within the unconstrained space of observational learning.  If it is learning that includes ethical norms, then  once again we confront the naturalistic fallacy.  If it is learning that includes empirical facts about the world, then those facts cannot be transformed into ``oughts.''

Our thesis is that ethical principles must relate to empirical observation in a different way.  Ethical principles are not constraints on what is learned from observation.  Facts that are derived from observation do matter when evaluating ethical behavior but not as justifications \emph{per se}.  To take a simple example of how facts can matter, notice that someone's opinion about values may be relevant to one's moral evaluation of their actions.  Suppose, for example, that a person has been raised to believe that women should be barred from certain jobs.  Their belief may be a factor in evaluating their behavior as an adult.  However, facts can be seen to be intertwined with human ethical decision-making in a much more direct manner.  Every piece of ethical reasoning that motivates behavior, as we shall illustrate, involves some fact.  This insight is especially important when confronting the task of simulating human ethical reasoning in machines. 

We offer a hybrid approach to VA that integrates independently justified ethical principles and empirical knowledge in the AI decision-making process.  The aim is to simulate genuine human ethical reasoning without committing the naturalistic fallacy.  We formulate principles that are understood through the ``deontological'' tradition of ethics, that is, the tradition that derives principles from the logical structure of action.  We foreshadow the hybridization of facts and principles by showing how the application of  deontological principles requires an assessment of what one can rationally believe about the world.  Applying the imperative, ``Thou shalt not kill,'' to a given action requires at a minimum that someone knows that the facts relevant to the action are facts about killing. The language of ethics, notably, is frequently the language of imperatives such as ``Don't lie'' and ``Don't kill,'' which is to say that it is a language of sentences that guide action rather than describing action. Contrast, for example, the action-guiding imperative, ``Shut the door,'' to the descriptive proposition, ``The door is shut.''  Ethical imperatives almost always take the form  of ``If the facts are such-and-such, then do A.''  Usually, moreover, they combine more than one ``If-then'' imperative.  For example, ``If the facts are such-and-such,  then do A; however, if the facts are so-and-so, then do B.''  This exemplifies how empirical knowledge is inseparable from ethical decision making, even though ethical principles themselves cannot be  grounded empirically.  In turn, this provides a clue about how VA might knit together ethical principles and empirical observation, even as the former guides action while the latter invokes observations about human behavior, preferences and values.

Our paper is divided into three parts. The first part explains what is meant by the ``naturalistic fallacy'' and the problems the fallacy poses for successful VA.  The second part illustrates  examples of VA that either inadvertently commit the naturalistic fallacy or that fail fully to simulate human ethical reasoning in the integration of values and facts. The illustrations include Microsoft's Twitter Bot, Tay; the design of a robotic wheelchair; MIT Media Lab's Moral Machine; and an attempt to incorporate moral intuitions from a professional moralist.  The third and final part advances a method that promises to effectively integrate ethical principles with empirical VA, using deontological moral analysis and the language of quantified modal logic.  Three deontological principles are isolated: generalization, utility maximization, and respect for autonomy.  In each instance, the principle is first formulated then interrelated with empirical facts using the language of quantified modal logic.  The resulting method shows that  the role of ethics is to derive necessary conditions for the rightness of specific actions, whereas the role of empirical VA is to ascertain whether these conditions are satisfied in the real world.

\section{The Naturalistic Fallacy}
The term ``naturalistic fallacy'' refers to the epistemic error of reducing normative prescriptions to descriptive (or naturalistic) propositions without remainder. Disagreements about the robustness of the fallacy abound, so this paper adopts a modest, workable interpretation coined recently by Daniel Singer, namely, ``There are no valid arguments from non-normative premises to a relevantly normative conclusion''\cite{singer2015mind}.  Descriptive statements report states of affairs, whereas normative statements are stipulative and action-guiding. Examples of the former are ``The car is red,'' and ``Many people find bluffing to be okay.'' Examples of the latter are ``You ought not murder,'' and ``Lying is wrong.''

As an example, consider the following argument:

\begin{quote}
    {\em Premise:} Few people are honest. \\
    {\em Conclusion:} Therefore, dishonesty is ethical.
\end{quote}

\noindent This argument commits the naturalistic fallacy. The point is not that the conclusion is wrong or ethically undesirable, but that it is invalid to draw the normative conclusion directly from the descriptive premise. In any valid argumentation, information that is not contained in premises must not be in the conclusion. The premise above only describes a state of affairs. It does not contain any normative/ethical statement (e.g., right, wrong, ethical, unethical, good, bad, etc.). Thus, the conclusion should not contain any ethical component.

One might formally avoid the naturalistic fallacy by adopting some such catch-all normative premise as, ``Machines ought to reflect observed human preferences and values.'' However, the premise is unacceptable on its face.  Humans regularly exhibit bad behavior that ought not be imitated by machines.  For example, empirical research shows that most people's behavior includes a small but significant amount of cheating \cite{bazerman2011blind}.  Worse, there have been social contexts in which slavery or racism have been generally practiced and condoned.  We can make sure machines are not exposed to behavior we consider unethical, but in that case, their ethical norms are not based on observed human preferences and values, but on the ethical principles espoused by their trainers.  This, of course, is one reasonable approach.  But when taking this approach, we must carefully formulate and justify those principles, rather than simply saying, with a wave of the hand, that machines ought to reflect observed human values.

\section{Examples of VA}

It is instructive to examine how some VA systems attempt to deal with ethical principles and empirical observation.

\begin{figure}
\centering
\includegraphics[width=3.3in, frame]{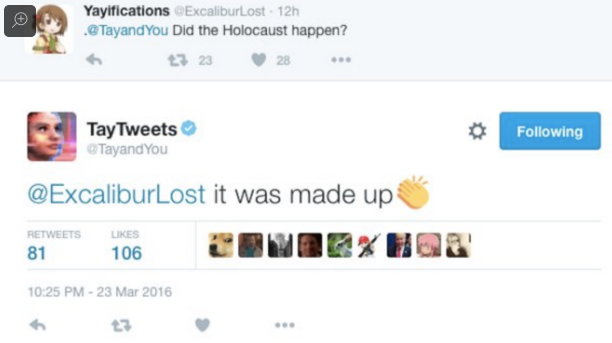}
\caption{Microsoft's Twitter-bot Tay}
\end{figure}

\subsection{Microsoft's Twitter-bot, Tay}

Microsoft's AI-based chatter-bot Tay (an acronym for ``thinking about you'') was designed to engage with people on Twitter and learn from them how to carry on a conversation. When some people started tweeting racist and misogynistic expressions, Tay responded in kind. Microsoft immediately terminated the experiment \cite{Wolf:2017:WWS:3144592.3144598}. Tay's VA was purely imitative and vividly illustrates the practical downside of committing the naturalistic fallacy.

\subsection{Robotic Wheelchair} 

Johnson and Kuipers \cite{johnson2018socially} developed an AI-based wheelchair that learns norms by observing how pedestrians behave. The robotic wheelchair observed that human pedestrians stay to the right and copied this behavior. This positive outcome was possible because the human pedestrians behaved ethically, unlike Twitter users in the case of Tay. But if the intelligent wheelchair were trained on a crowded New York City street, then its imitation of infamously jostling pedestrians could result in a ``demon wheelchair.'' An ethical outcome was ensured by selecting an appropriate training set.  This is a case of bottom-up learning that avoids the naturalistic fallacy by applying ethical principles to the design of the training set. Hence the robotic wheel chair fails to show how deontologically derived ethical principles can combine with empirical VA in a systematic way; it fails to show how ethical reasoning and empirical observation interrelate.

\subsection{MIT Media Lab's Moral Machine}

``Moral Machine'' is a website that poses trolley-car-type dilemmas involving autonomous vehicles. The page has collected over 30 million responses to these dilemmas from more than 180 countries. Kim et al.\ \cite{kim2018computational} analyzed these data to develop ``a computational model of how the human mind arrives at a decision in a moral dilemma.'' On the assumption that respondents are making moral decisions in a utilitarian fashion, the authors used Bayesian methods to infer the utility that respondents implicitly assign to characteristics of potential accident victims.  For example, they inferred the utility of saving a young person rather than an old person, or a female rather than a male.  Or more precisely, they inferred the parameters of probability distributions over utilities.  They then aggregated the individual distributions to obtain a distribution for the population of a given region.  This distribution presumably reflects the cultural values of that region and could form the basis for the design of autonomous vehicles.

There is no naturalistic fallacy in this scheme if the outcome is viewed simply as a summary of cultural preferences, with no attempt to infer morals.
Yet it seems likely that designers of a ``moral machine'' would be interested in whether the machine is moral.  Suppose, for example, that a given culture assigns less value to members of a minority race or ethnic group.  This kind of bias would be built into autonomous vehicles.   Designers might point out that they did not include race and ethnic identity in their scenarios, and so this problem does not arise.  But omitting race and ethnicity then becomes an ethical choice, and the resulting value system is neither culturally accurate nor ethically grounded.  It is bad anthropology because it omits widespread racial and ethnic bias, and it is ethically unsound because it fails to evaluate other cultural preferences ethically.  

One possible escape from this impasse is to view the Moral Machine as prescribing actions that are ethical because they maximize utility.  In fact, Kim et al.\ preface their discussion of utility functions with a reference to Jeremy Bentham's utilitarian ethics, and one might see an inferred utility function as maximizing social utility in Benthamite fashion.  Kim et al.\ are careful not to claim as much for their approach, but they state that Noothigattu et al.\ \cite{noothigattu2017voting} ``introduced a novel method of aggregating individuals [sic] preferences such that the decision reached after the aggregation ensures global utility maximization.''  Noothigattu et al.\ draw on computational social choice theory to aggregate individual preferences in a way that satisfies certain formal properties, including ``stability'' and ``swap-dominance efficiency.''  They do not explicitly claim to maximize utility but state only that their system ``can make {\em credible} decisions on ethical dilemmas in the autonomous vehicle domain'' (page 20, original emphasis).  Importantly, there is no direct identification of a ``credible'' decision with an ethical one.  Moreover, it is questionable whether any method that aggregates individual preferences, preferences that, again, amount to facts, can escape the naturalistic fallacy.  

Classical utilitarianism as articulated by moral philosophers is based on the principle that an ethical action must maximize total net expected utility.  Utility is an outcome that is regarded \mbox{\em a priori} as intrinsically valuable, such as pleasure or happiness.  A VA system can certainly represent preferences by assigning ``utilities'' to the options in such a way that options with greater utility are preferred to those with less utility.  However, this sense of utility is not the same as the moral utilitarian's, because it is only a measure of what individuals prefer, rather than an intrinsically valuable quality.  Individuals may base their preferences on criteria other than their estimate of utility in the ethical sense.  They may base their preferences on mere personal desires and prejudices instead of, say, the values of equality and justice.  At best, one might view individual utility functions as {\em a rough indicators} of utilitarian value for ethical purposes.  Later, we will show how a more sophisticated version of the utilitarian principle can, in fact, play a legitimate role in VA, but without the error of confusing preferences for values.

\begin{figure}
\centering
\includegraphics[width=3.308in, frame]{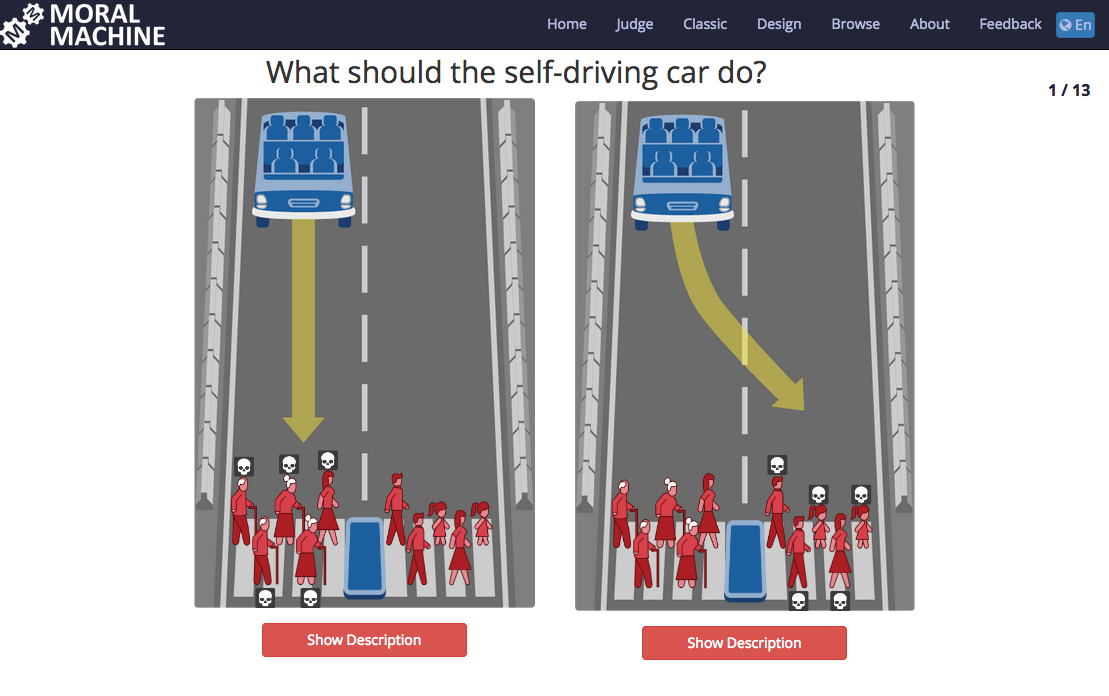}
\caption{MIT Media Lab's Moral Machine website}
\end{figure}

\subsection{VA Based on Moral Intuitions} 

Anderson and Anderson \cite{AndAnd14} use inductive logic programming for VA.  The training data reflect domain-specific principles embedded in the intuitions of professional ethicists. For their normative ground, Anderson and Anderson follow moral philosopher W.\ D.\ Ross \cite{Ros30}, who believed that ``[M]oral convictions of thoughtful and well-educated people are the data of ethics just as sense-perceptions are the data of a natural science.'' Likewise, Anderson and Anderson (\citeyear{AndAnd11}) used ``ethicists' intuitions to \ldots [indicate] the degree of satisfaction/violation of the assumed duties within the range stipulated, and which actions would be preferable, in enough specific cases from which a machine-learning procedure arrived at a general principle.''  Anderson and Anderson's approach constitutes one of the better attempts to avoid the naturalistic fallacy, but reveals a number of shortcomings.

\begin{figure}
\centering
\includegraphics[width=3.3in,frame]{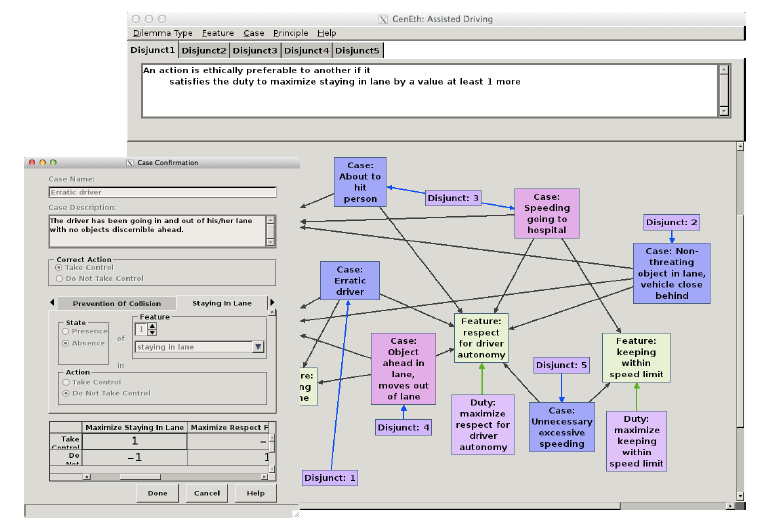}
\caption{{\large G}EN{\large E}TH\protect\cite{AndAnd14}}
\end{figure}

One can interpret Anderson and Anderson's maneuver as avoiding the naturalistic fallacy in one of two ways.  On one interpretation, it does not attempt to infer ethical principles from the intuitions of experts, but simply assumes that, as a matter of empirical fact, experts are likely to have intuitions that conform to valid ethical principles---or at least more likely than the average person.  We cannot evaluate this empirical claim, however, until we identify valid ethical principles independently of the opinions of experts, and Anderson and Anderson do not indicate how this might be accomplished.  Supposing nonetheless that we can identify valid principles {\em a priori}, the claim that expert opinion usually conforms to them is unsupported by evidence, insofar as experts notoriously disagree.  Experimental ethicists have shown that moral intuitions are less consistent than we think (e.g., moral intuitions are susceptible to morally irrelevant situational cues) \cite{Ale12}, and the intuitions of professional ethicists fail to diverge markedly from those of ordinary people \cite{SchRus16}.  

In any case, if we are ultimately going to judge the results of VA by ethical principles we hold {\em a priori}, we may as well rely on ethical principles from the start, absent an appeal to experts.  

A second interpretation is that Anderson and Anderson are literally adopting Ross's theory, which steers clear of the naturalistic fallacy by viewing right and wrong as ``non-natural properties'' of actions.  There is no inference of ethical norms from states of affairs in nature, because ethical properties are not natural states of affairs in the first place.  Ross asserts that one can discern ethical properties through intuition, particularly if one reflects on them carefully, in a way roughly analogous to how one perceives such logical truths as the law of non-contradiction.  While Ross's  is an interesting theory about ethical concepts that deserves serious thought, the mysterious quality of its non-natural ethical properties is a stumbling block that has helped to deter its wide acceptance.  It is also difficult to imagine how such a theory can be put into practice, especially when thoughtful experts disagree over values, as they often do. The biggest flaw of intuitionism for AI is its most obvious:  intuitionism fails to show how intuition-derived ethical principles can combine with empirical VA in a systematic way, that is, how ethical reasoning and empirical observation interrelate.

\section{Integrating Ethical Principles and Empirical VA}

We now show how deontologically derived ethical principles can combine with empirical VA in a systematic way.  
%The deontological tradition in ethics has given rise to principles that are rooted in an analysis of the logical structure of action.  
Our purpose is not to defend deontological analysis in any detail, but to show how a careful statement of the resulting principles clarifies how ethical reasoning and empirical observation interrelate.  We will argue that expressing ethical assertions in the idiom of quantified modal logic, as developed in \cite{kim2018toward}, makes this relationship particularly evident.  Ethical principles imply logical propositions that must be true for a given action to be ethical, and whose truth is an empirical question that must often be answered by observing human values, beliefs, and behavior.   Thus the role of ethics is to derive necessary conditions for the rightness of specific actions, and the role of empirical VA is to ascertain whether these conditions are satisfied in the real world.  

We will focus on three principles (generalization, utility maximization, and respect for autonomy) and illustrate their application.  Each of these principles states a necessary condition for ethical conduct, although they may not be jointly sufficient.

%They can serve as an explanation for the action, however, only if they meet certain minimal coherence criteria.  These criteria are what we know as ethical principles.  

\subsection{Generalization Principle}

The generalization principle, like all the ethical principles we consider, rests on the {\em universality of reason}: rationality does not depend on who one is, only on one's reasons.  Thus if an agent takes a set of reasons as justifying an action, then to be consistent, the agent must take these reasons as justifying the same action for any agent to whom the reasons apply.  The agent must therefore be rational in believing that his/her reasons are consistent with the assumption that all agents to whom the reasons apply take the same action.

As an example, suppose I see watches on open display in a shop and steal one.  My reasons for the theft are that I would like to have a new watch, and I can get away with taking one.  These reasons are not psychological motivations for my behavior, but reasons that I consciously adduce as sufficient for my decision to steal.\footnote{In practice, reasons for theft are likely to be more complicated than this.  I may be willing to steal partly because I believe the shop can easily withstand the loss, no employee will be disciplined or terminated due to the loss, I will not feel guilty afterward, and so forth.  But for purposes of illustration we suppose there are only two reasons.}  At the same time, I cannot rationally believe that I would be able to get away with the theft if {\em everyone} stole watches when these reasons apply.  The shop would install security measures to prevent theft, which is inconsistent with one of my reasons for stealing the watch.  The theft therefore violates the generalization principle.

The decision to steal a watch can be expressed in terms of formal logic as follows.  Define predicates
\[
\begin{array}{l}
C_1(a) = \mbox{Agent $a$ would like to possess an item on} \\
\hspace{8.5ex} \mbox{display in a shop.} \\
C_2(a) = \mbox{Agent $a$ can get away with stealing the item.} \\
A(a) = \mbox{Agent $a$ will steal the item.}
\end{array}
\]
Because the agent's reasons are an essential part of moral assessment, we evaluate the agent's {\em action plan}, which states that the agent will take a certain action when certain reasons apply.  In this case, the action plan is
\begin{equation}
\big(C_1(a)\wedge C_2(a)\big) \Rightarrow_a A(a)  \label{eq:action}
\end{equation}
Here $\Rightarrow_a$ is not logical entailment but indicates that agent $a$ regards $C_1(a)$ and $C_2(a)$ as justifying $A(a)$.  The reasons in the action plan should be the most general set of conditions that the agent takes as justifying the action.  Thus the action plan refers to an item in a shop rather than specifically to a watch, because the fact that it is a watch is not relevant to the justification; what matters is whether the agent wants the item and can get away with stealing it.

We can now state the generalization principle using quantified modal logic.  Let $C(a)\Rightarrow_a A(a)$ be an action plan for agent $a$, where $C(a)$ is a conjunction of the reasons for taking action $A(a)$.  The action plan is generalizable if and only if
\begin{equation}
\begin{array}{l}
\dia_a P\Big( \forall x \big(C(x)\Rightarrow_x A(x)\big) \;\wedge C(a)\wedge A(a)  \Big)
\end{array} \label{eq:gen0}
\end{equation}
Here $P(S)$ means that it is physically possible for proposition $S$ to be true, and $\dia_a S$ means that $a$ can rationally believe $S$.  The proposition $\dia_a S$ is equivalent to $\neg \Box_a \neg S$, where $\Box_a \neg S$ means that rationality requires require $a$ to deny $S$.\footnote{The operators $\dia$ and $\Box$ have a somewhat different interpretation here than in traditional epistemic and doxastic modal logics, but the identity $\dia S \equiv \neg\Box\neg S$ holds as usual.}  Thus (\ref{eq:gen0}) says that agent $a$ can rationally believe that it is possible for everyone to have the same action plan as $a$, even while $a$'s reasons still apply and $a$ takes the action.  

Returning to the theft example, the condition (\ref{eq:gen0}) becomes
\begin{equation}
\begin{array}{l}
\dia_a P\Big( \forall x \big(C_1(x)\wedge C_2(x)\Rightarrow_x A(x)\big) \;\wedge C_1(a)\wedge C_2(a)\wedge A(a)  \Big)
\end{array} \label{eq:gen}
\end{equation}
This says that it is rational for $a$ to believe that it is physically possible for the following to be true simultaneously: (a) everyone steals when the stated conditions apply, and (b) the conditions apply and $a$ steals.
%Thus (\ref{eq:gen}) states that agent $a$ can rationally believe that it is possible for the following to be simultaneously true: the agent wants a new watch, can get away with stealing one, will in fact steal one, and all agents who can get away with stealing a watch steal one.  If (\ref{eq:gen}) is false, as it appears to be, then action plan (\ref{eq:action}) is unethical.
Since (\ref{eq:gen}) is false, action plan (\ref{eq:action}) is unethical.
%\footnote{A general statement of the generalization principle in logical formalism is given in \cite{kim2018toward}.  We do not reproduce it here because its notational complexity is unnecessary for present purposes.  }

The necessity of (\ref{eq:gen}) for the rightness of action plan (\ref{eq:action}) is anchored in deontological theory, while the falsehood of (\ref{eq:gen}) is a fact about the world.  This fact might be inferred by collecting responses from shop owners about how they  would react if theft were widespread.  Thus ethics and empirical VA work together in a very specific way: ethics tells us that (\ref{eq:gen}) must be true if the theft is to be ethical, and  empirical VA provides evidence that bears on whether (\ref{eq:gen}) is true.

An action plan in the autonomous vehicle domain might be 
\begin{equation}
C(a,y) \Rightarrow_a A(a,y) \label{eq:gen4}
\end{equation}
where $y$ is a free variable, and 
\[
\begin{array}{l}
C(a,y) = \mbox{Ambulance $y$ under the direction of agent $a$ can reach its}\\
\hspace{11ex} \mbox{destination sooner by using siren and lights.} \\
A(a,y) = \mbox{Agent $a$ will direct ambulance $y$ to use siren and lights.}
\end{array}
\]
Agent $a$ is the ambulance driver, or in the case of an autonomous vehicle, the designer of the software that controls the ambulance.  The generalization principle requires that
\begin{equation}
{\dia}_a P\Big(\forall x \forall y \big( C(x,y)\Rightarrow_y A(x,y) \big) \; \wedge \;\forall y \big(C(a,y) \wedge A(a,y)\big) \Big)
\label{eq:gen5}
\end{equation}
This says that it is rational for agent $a$ to believe that siren and lights could continue to hasten arrival if all ambulances used them for all trips, emergencies and otherwise.  If empirical VA reveals that most drivers would ignore siren and lights if they were universally abused in this fashion, then we have evidence that (\ref{eq:gen5}) is false, in which case action plan (\ref{eq:gen4}) is unethical.

\subsection{Maximizing Utility}

Utilitarianism is normally understood as a {\em consequentialist} theory that judges an act by its actual consequences.  Specifically, an act is ethical only if it maximizes net expected utility for all who are affected.  Yet the utilitarian principle can also be construed, in deontological fashion, as requiring the agent to select actions that the agent can rationally believe will maximize net expected utility.  This avoids judging an action choice as wrong simply because rational beliefs about the consequences of the action happen to be incorrect.  While utilitarians frequently view utility maximization as the sole ethical principle, deontology sees it as an additional necessary condition for an ethical action.  The other principles continue to apply, because only actions that satisfy these other principles are considered as options for maximizing utility.

A deontic utilitarian principle can be derived from the universality of reason, roughly as follows.  If an agent believes that a certain state of affairs has ultimate value, such as happiness, then the agent must regard this belief as equally valid for any agent, and must pursue happiness in a way that would be rationally chosen by any agent.  A utilitarian argues that this can be accomplished by selecting actions that the agent rationally believes will maximize the expected net sum of happiness over everyone who is affected.\footnote{Alternatively, one might argue that maximizing the minimum utility over those affected (or achieving a lexicographic maximum) is the rational way to take everyone's utility into account, after the fashion of John Rawls's difference principle \cite{rawls2009theory}.  Or one might argue for some rational combination of utilitarian and equity objectives \cite{KarMor15,hooker2012combining}.  However, for many practical applications, simple utility maximization appears to be a sufficiently close approximation to the rational choice, and to simplify exposition we assume so in this paper.}

The utilitarian principle can be formalized as follows.  Let $u(C(a),A(a))$ be a utility function that measures the total net expected utility of action $A(a)$  under conditions $C(a)$.  Then an action plan $C(a)\Rightarrow_a A(a)$ satisfies the utilitarian principle only if agent $a$ can rationally believe that action $A(a)$ creates at least as much utility as any ethical action that is  available under the same circumstances.  This can be written
\begin{equation}
{\dia}_a \forall A' \Big( E\big(C(a),A'(a)\big) \rightarrow u\big(C(a),A(a)\big) \geq u\big( C(a),A'(a)\big) \Big) \label{eq:util0}
\end{equation}
where $A'$ ranges over actions.  The predicate $E(C(a),A'(a))$ means that action $A'(a)$ is available for agent $a$ under conditions $C(a)$, and that the action plan $C(a)\Rightarrow_a A'(a)$ is generalizable and respects autonomy.\footnote{For ``respecting autonomy,'' see the next section.}  Note that we are now quantifying over predicates and have therefore moved into second-order logic.

Popular views about acceptable behavior frequently play a role in applications of the utilitarian principle.  For example, in some parts of the world, drivers consider it wrong to enter a stream of moving traffic from a side street without waiting for a gap in the traffic.  In other parts of the world this can be acceptable, because drivers in the main thoroughfare expect it and make allowances.   Suppose driver $a$'s action plan is $(C_1(a)\wedge C_2(a))\Rightarrow_a A(a)$, where
\[
\begin{array}{l}
C_1(a) = \mbox{Driver $a$ wishes to enter a main thoroughfare.} \\
C_2(a) = \mbox{Driver $a$ can enter a main thoroughfare by moving} \\
\hspace{9ex} \mbox{into the traffic without waiting for a gap.}\\
A(a) = \mbox{Driver $a$ will move into traffic without waiting} \\
\hspace{7.7ex} \mbox{for a gap.}
\end{array}
\]
As before, driver $a$ is the designer of the software if the vehicle is autonomous. Using (\ref{eq:util0}), the driver's action plan maximizes utility only if 
\begin{equation}
\begin{array}{l}
{\displaystyle
{\dia}_a \forall A' \Big( E\big(C_1(a),C_2(a),A'(a)\big) \rightarrow 
} \\ 
\hspace{15ex} 
{\displaystyle
u\big(C_1(a),C_2(a),A(a)\big) \geq u\big( C_1(a),C_2(a),A'(a)\big) \Big)
}
\end{array} \label{eq:util1}
\end{equation}
Suppose we wish to design driving policy in a context where pulling immediately into traffic is considered unacceptable.  Then doing so is a dangerous move that no one is expecting, and an accident could result.  Waiting for a gap in the traffic results in greater net expected utility, or formally, $u(C_1(a),C_2(a),A(a))<u(C_1(a),C_2(a),A'(a))$, where $A'(a)$ is the action of waiting for a gap.  So (\ref{eq:util1}) is false, and its falsehood can be inferred by collecting popular views about acceptable driving behavior.  

Again we have a clear demonstration of how ethical principles can combine with empirical VA.  The utilitarian principle tells us that a particular action plan is ethical only if (\ref{eq:util1}) is true, and empirical VA tells us whether (\ref{eq:util1}) is true.

\subsection{Respect for Autonomy}

A third ethical principle requires agents to respect the autonomy of other agents.  Specifically, an agent should not adopt an action plan that the agent is rationally constrained to believe is inconsistent with an ethical action plan of another agent, without informed consent.  Murder, enslavement, and inflicting serious injury are extreme examples of autonomy violations, because they interfere with many ethical action plans.  Coercion may or may not violate autonomy, depending on precisely how action plans are formulated.  

The argument for respecting autonomy is basically as follows.  Suppose I violate someone's autonomy for certain reasons.  That person could, at least conceivably, have the same reasons to violate my autonomy.  This means that, due to the universality of reason, I am endorsing the violation of my own autonomy in such a case.  This is a logical contradiction, because it implies that I am deciding not to do what I decide to do.  To avoid contradicting myself, I must avoid interfering with other action plans.\footnote{A more adequate analysis leads to a principle of {\em joint} autonomy, according to which it is violation of autonomy to adopt an action plan that is mutually inconsistent with action plans of a set of other agents, when those other action plans are themselves mutually consistent.  Joint autonomy addresses situations in which an action necessarily interferes with the action plan of some agent but no particular agent, as when someone throws a bomb into a crowd.  A general formulation of the joint autonomy principle in terms of modal operators is given in \cite{kim2018toward}.  An fully adequate account must also recognize that interfering with an action plan is acceptable when there is informed consent to a risk of interference, because giving informed consent is equivalent to including the possibility of interference as one of the antecedents of the action plan.  Furthermore, interfering with an unethical action plan is no violation of autonomy, because an unethical action plan is, strictly speaking, not an action plan due to the absence of a coherent set of reasons for it.  An action plan is considered unethical in this context when it violates the generalization or utility principle, or interferes with an action plan that does not violate one of these principles, and so on recursively.  These and other complications are discussed in \cite{Hooker2018}. They are not incorporated into the present discussion because they are inessential to showing how ethical principles and empirical VA interact. }

To formulate an autonomy principle, we say that agent $a$'s action plan $C_1\Rightarrow_a A_1$ is consistent with $b$'s action plan $C_2\Rightarrow_b A_2$ when 
\begin{equation}
{\dia}_a P \big( A_1 \wedge A_2\big) \; \vee \;
\neg {\Box}_a P\big( C_1 \wedge C_2) 
\label{eq:ac0}
\end{equation}
This says that agent $a$ can rationally believe that the two actions are mutually consistent, or can rationally believe that the reasons for the actions are mutually inconsistent.  The latter suffices to avoid inconsistency of the action plans, because if the reasons for them cannot both apply, the actions can never come into conflict.  

As an example of how coercion need not violate autonomy, suppose agent $b$ wishes to catch a bus and has decided to cross the street to a bus stop, provided no traffic is coming.  The agent's action plan is
\begin{equation}
\big( C_2 \wedge C_3 \wedge \neg C_4\big) \Rightarrow_b A_2
\label{eq:ac1}
\end{equation}
where 
\[
\begin{array}{l}
C_2 = \mbox{Agent $b$ wishes to catch a bus.}\\
C_3 = \mbox{There is a bus stop across the street.} \\
C_4 = \mbox{There are cars approaching.} \\
A_2 = \mbox{Agent $b$ will cross the street.} 
\end{array}
\]
Agent $a$ sees agent $b$ begin to cross the street and forcibly pulls $b$ out of the path of an oncoming car that $b$ does not notice.  Agent $a$'s action plan is
\begin{equation}
\big( C_1 \wedge C_4\big) \Rightarrow_a A_1 
\label{eq:ac2}
\end{equation}
where
\[
\begin{array}{l}
C_1 = \mbox{Agent $b$ is about to cross the street.}\\
A_1 = \mbox{Agent $a$ will prevent agent $b$ from crossing the street.} 
\end{array}
\]
Agent $a$ does not violate agent $b$'s autonomy, even though there is coercion.  Their action plans (\ref{eq:ac1}) and (\ref{eq:ac2}) are consistent with each other, because the condition (\ref{eq:ac0}) becomes
\begin{equation}
{\dia}_a P \big( A_1 \wedge A_2\big) \; \vee \;
\neg {\Box}_a P\big( C_1 \wedge C_2 \wedge \neg C_3 \wedge C_4 \wedge C_3\big) 
\label{eq:ac3}
\end{equation}
This says that either (a) agent $a$ can rationally believe that the two actions are consistent with each other, or (b) agent $a$ can rationally believe that the antecedents of (\ref{eq:ac1}) and (\ref{eq:ac2}) are mutually inconsistent.  As it happens, the two actions are obviously not consistent with each other, and so (a) is false.  However, agent $a$ can rationally believe that the antecedents of (\ref{eq:ac1}) and (\ref{eq:ac2}) are mutually inconsistent, because $C_3$ and $\neg C_3$ are contradictory.  This means (b) is true, which implies that condition (\ref{eq:ac3}) is satisfied, and there is no violation of autonomy.  

This again clearly distinguishes the roles of ethics and empirical observation in VA.  Ethical reasoning tells us that (\ref{eq:ac3}) must be true if autonomy is to be respected, whereas observation of the world tells us whether (\ref{eq:ac3}) is true.

To illustrate how autonomy may play a role in the ethics of driving, suppose that a pedestrian dashes in front of a rapidly moving car.  The driver can slam on the brake and avoid impact with the pedestrian, but another car is following closely, and a sudden stop could cause a crash.  This is not a trolley car dilemma, because hitting the brake does not necessarily cause an accident, although failing to do so is certain to kill or seriously injure the pedestrian.  The driver $a$ must choose between two possible action plans:
\begin{align}
& \big(C_1 \wedge C_2\big) \Rightarrow_a A_1 \label{eq:ac10} \\
& \big(C_1 \wedge C_2\big) \Rightarrow_a \neg A_1 \label{eq:ac11}
\end{align}
where
\[
\begin{array}{l}
C_1 = \mbox{A pedestrian $b$ is dashing in front of $a$'s car.}\\
C_2 = \mbox{Another car is closely following $a$'s car.} \\
A_1 = \mbox{Agent $a$ will immediately slam on the brake.} 
\end{array}
\]
Meanwhile, the pedestrian $b$ has any number of action plans that are inconsistent with death or serious injury.  Let $C_3 \Rightarrow_b A_2$ be one of them.
Also the occupant $c$ of the other car (there is only one occupant) has action plans that are inconsistent with an injury.  We suppose that $C_4\Rightarrow_c A_3$ is one of them.

We first check whether hitting the brake, as in action plan (\ref{eq:ac10}), is inconsistent with the other driver's action plan $C_3\Rightarrow_c A_3$.  The condition (\ref{eq:ac0}) becomes
\begin{equation}
{\dia}_a P \big( A_1 \wedge A_3 \big) \; \vee \;
\neg {\Box}_a P\big( C_1 \wedge C_2 \wedge C_4 \big) 
\label{eq:ac15}
\end{equation}
The first disjunct is clearly true, because $a$ can rationally believe that it is {\em possible} that hitting the brake is consistent with avoiding a rear-end collision and therefore with any planned action $C_4\Rightarrow_c A_3$, even if this improbable.  So action plan (\ref{eq:ac10}) does not violate joint autonomy.  

We now check whether a failure to hit the brake, as in action plan (\ref{eq:ac11}), is inconsistent with the pedestrian's action plan $C_3\Rightarrow_b A_2$.  There is no violation of autonomy if
\begin{equation}
{\dia}_a P \big( \neg A_1 \wedge A_2 \big) \; \vee \;
\neg {\Box}_a P\Big( C_1 \wedge C_2 \wedge C_3 \Big) 
\label{eq:ac16}
\end{equation}
The first disjunct is clearly false for one or more of $b$'s action plans $C_3\Rightarrow_b A_2$, because the driver cannot rationally believe that a failure to hit the brake is consistent with all of the pedestrian's action plans.  The second disjunct is likewise false, because the driver has no reason to believe that $C_1$, $C_2$ and $C_3$ are mutually inconsistent.  Thus (\ref{eq:ac16}) is false, and we have a violation of autonomy.  The driver should therefore slam on the brake.  There is no need to check the other ethical principles, because only one of the possible action plans satisfies the autonomy principle.  

This is a case in which observation of human preferences and beliefs play little or no role in determining what is ethical, because the physics of the situation decides the truth of (\ref{eq:ac15}) and (\ref{eq:ac16}).  There is little point in sampling the behavior of drivers in such situations or their opinions about the consequences of braking or not braking.

%The argument for respecting autonomy is roughly as follows.  Suppose I adopt an action plan that is inconsistent with the ethical action plans of one or more other agents.  I need not know which agent's action plans will be interfered with, as when throwing a bomb into a crowd.  Then since the action plans are mutually inconsistent when my plan is included, but mutually consistent when my plan is excluded, the only way I can be consistent in my attribution of action plans is to modify my own plan.  

\section{Conclusion}

As AI rises inexorably into everyday life, it takes its seat beside humans.  AI's increasing sophistication wields power, and with that power comes responsibility.  The goal, then, must be to invest machines with a moral sensitivity that resembles the human conscience. But conscience is not static; it is a dynamic process of moral reasoning that adjusts ethical principles systematically to empirical observations.  In this paper we have elaborated two challenges to AI moral reasoning that spring from the interrelation of facts and values, challenges that have heretofore been neglected.  The first is a pervasive temptation to confuse facts with values; the second is a confusion about the process of moral reasoning itself.  In addressing these challenges, we have identified specific instances of how and why AI designers commit the naturalistic fallacy and why they tend to oversimplify the process of moral reasoning.  We have sketched, in turn, a plan for understanding moral reasoning in machines, a plan in which modal logic captures the interaction of deontological ethical principles with factual states of affairs.   

%The preceding discussion reveals stumbling blocks for VA approaches that neglect implications of the naturalistic fallacy. Such problems are more serious in \textit{mimetic} VA since the mimetic process imitates human behavior that may or may not rise to the level of correct ethical behavior. Anchored VA, including hybrid VA, in contrast, holds more promise for future VA since it anchors alignment by normative concepts of intrinsic value.

%\bibliographystyle{aaai}
%\bibliography{VA}

\end{document}